\documentclass[10pt, conference]{IEEEtran}
\usepackage{orcidlink}

\usepackage{amsmath,amssymb,amsfonts}
\usepackage{pifont}

\usepackage{amsthm}
\usepackage{xcolor}
\usepackage[figuresright]{rotating}

\usepackage{graphicx}
\graphicspath{{/}{fig/}}

\usepackage{array}
\usepackage{textcomp}
\usepackage{multirow}
\usepackage{booktabs}

\usepackage{mathtools}
\usepackage{breqn}
\usepackage{float}

\usepackage{pgfplots}
\pgfplotsset{compat=1.7}
\usepgfplotslibrary{groupplots}

\usepackage{caption}
\usepackage{subcaption}

\usepackage{tabularx}
\usepackage{hyperref}
\usepackage{flushend}
\usepackage{algorithmic}
\usepackage[vlined, ruled, shortend]{algorithm2e}

\usepackage{enumitem}
\usepackage{gensymb}

\newlength\figureheight
\newlength\figurewidth
\SetAlCapNameFnt{\footnotesize}
\SetAlCapFnt{\footnotesize}

\title{
    Towards Robust UAV Tracking in GNSS-Denied Environments: A Multi-LiDAR Multi-UAV 
    Dataset 
}

\author{
    \IEEEauthorblockN{
        \vspace{1em}
        Iacopo Catalano\IEEEauthorrefmark{2}\,\orcidlink{0000-0001-9212-8615},
        Xianjia Yu\IEEEauthorrefmark{2}\,\orcidlink{0000-0002-9042-3730},
        Jorge Peña Queralta\IEEEauthorrefmark{2}$^{,}$\IEEEauthorrefmark{3}\,\orcidlink{0000-0003-3091-3217}
    }
    \IEEEauthorblockA{
        \normalsize
        \IEEEauthorrefmark{2}\href{https://tiers.utu.fi}{Turku Intelligent Embedded and Robotic Systems (TIERS) Lab, University of Turku, Finland}.\\
        \IEEEauthorrefmark{3}\href{https://scai.ethz.ch/}{SCAI Laboratory at SPZ, Swiss Federal School of Technology in Zurich - ETH Zurich, Switzerland}.\\
        Emails: \textsuperscript{1}\{iacopo.catalano, xianjia.yu\}@utu.fi, \textsuperscript{1}\{jorge.penaqueralta\}@hest.ethz.ch\\[+6pt]
    }
}

\begin{document}

\maketitle
\thispagestyle{empty}
\pagestyle{empty}



\begin{abstract}%
    \label{sec:abstract}%
    With the increasing prevalence of drones in various industries, the navigation and tracking of unmanned aerial vehicles (UAVs) in challenging environments, particularly GNSS-denied areas, have become crucial concerns. To address this need, we present a novel multi-LiDAR dataset specifically designed for UAV tracking. Our dataset includes data from a spinning LiDAR, two solid-state LiDARs with different Field of View (FoV) and scan patterns, and an RGB-D camera. This diverse sensor suite allows for research on new challenges in the field, including limited FoV adaptability and multi-modality data processing.

    The dataset facilitates the evaluation of existing algorithms and the development of new ones, paving the way for advances in UAV tracking techniques. Notably, we provide data in both indoor and outdoor environments. We also consider variable UAV sizes, from micro-aerial vehicles to more standard commercial UAV platforms. The outdoor trajectories are selected with close proximity to buildings, targeting research in UAV detection in urban areas, e.g., within counter-UAV systems or docking for UAV logistics.

    In addition to the dataset, we provide a baseline comparison with recent LiDAR-based UAV tracking algorithms, benchmarking the performance with different sensors, UAVs, and algorithms. Importantly, our dataset shows that current methods have shortcomings and are unable to track UAVs consistently across different scenarios.

    The dataset is available on GitHub\footnote{\url{https://github.com/TIERS/uav\_multi\_lidar\_dataset}}.
\end{abstract}

\IEEEpeerreviewmaketitle


\section{Introduction}\label{sec:introduction}

Unmanned Aerial Vehicles (UAVs) are gaining widespread use across diverse application domains due to their agility and ease of deployment~\cite{tsouros2019review, wang2019surveying}. Equipped with only a flight controller and basic sensor suite, they serve as efficient and adaptable mobile sensing platforms~\cite{gawel2018aerial, nex2014uav}. Recent research has focused on UAV navigation in GNSS-denied environments ~\cite{nieuwenhuisen2016autonomous, sier2023uav, catalano2023uav, jiang2021anti}, as well as state estimation in both single and multi-UAV systems~\cite{queralta2020collaborative, queralta2020autosos}.

The integration of UAVs into multi-robot systems emphasizes the importance of tracking between robots for relative or global state estimation methods~\cite{queralta2022vio, bai2020cooperative}. Tracking UAVs from an unmanned ground vehicle (UGV) within multi-robot setups allows for miniaturization and enhanced flexibility, reducing the reliance on high-accuracy onboard localization~\cite{petrlik2020robust, rouvcek2019darpa}. The UGV often acts as a base station, supplying crucial data to enable UAV operation even in areas with limited GNSS signals.

\begin{figure}
    \centering
    \includegraphics[width=.49\textwidth]{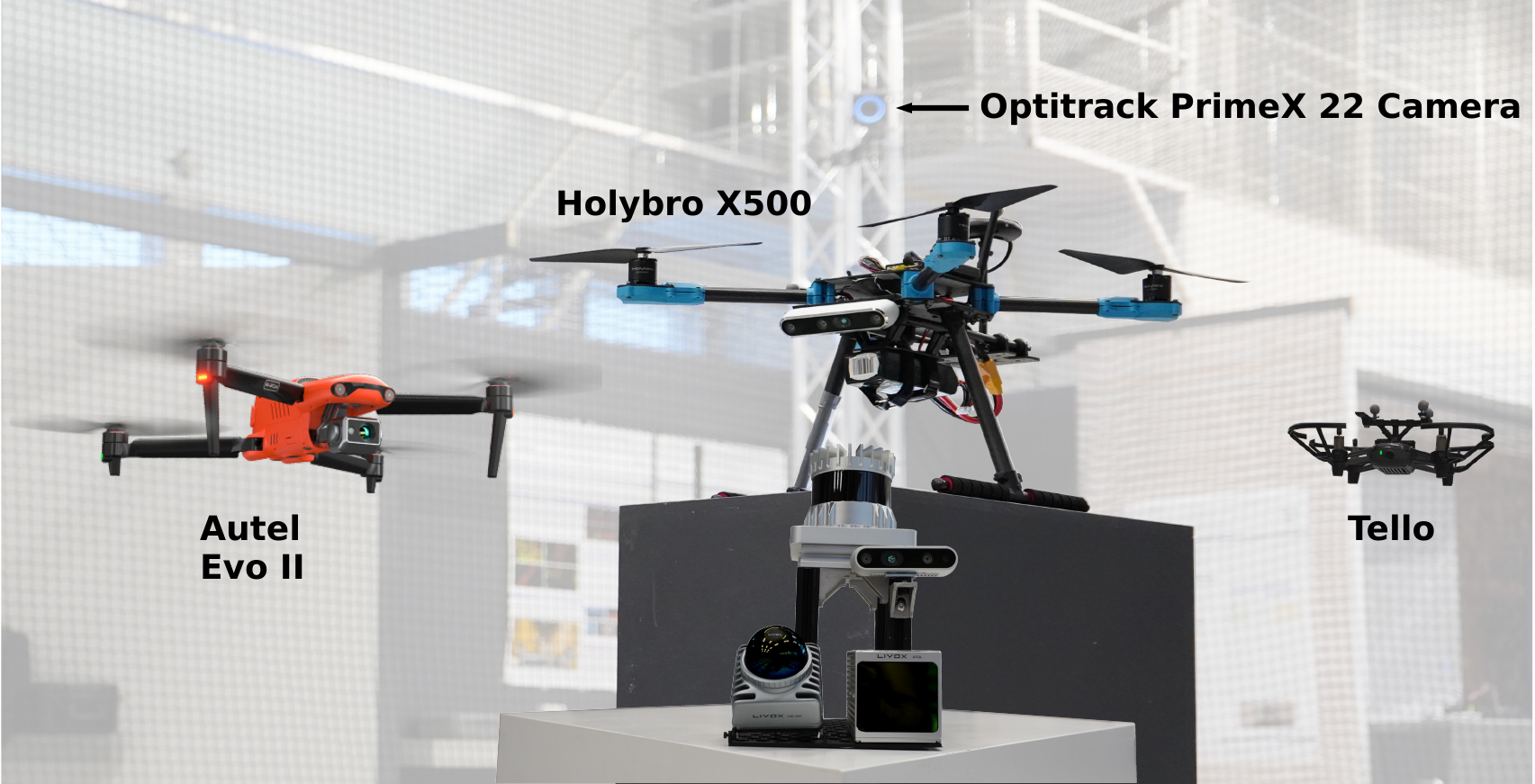}
    \caption{Illustration of the hardware used in the experiments. At the bottom, the tracking sensors including Ouster OS1-64, Livox Mid-360, Livox Avia and Intel RealSense D435.}
    \label{fig:hardware}
\end{figure}

Despite significant progress in UAV tracking using GNSS and other sensors, challenges persist. GNSS signal unavailability in certain environments, like indoors or urban canyons, restricts the accuracy and reliability of UAV positioning and tracking. Additionally, existing methods may rely on costly hardware or necessitate substantial computational power, affecting their practicality and scalability~\cite{li2020towards, osco2021review}. These limitations and challenges underscore the need for new approaches that are more robust, accurate, and efficient, capable of operating effectively in GNSS-denied environments. Moreover, external tracking of UAVs from the ground has gained relevance in the context of counter-UAV solutions~\cite{samaras2019deep}.

One of our primary objectives is to address the scarcity of data from solid-state LiDARs. These cutting-edge sensors, a recent development in long-range scanning technology, produce high-density point clouds, making them highly suitable for tracking objects in three-dimensional space, including UAVs~\cite{li2020towards}. Their non-repetitive scan patterns allow for the generation of dense point clouds with adjustable frequencies and variable field of view (FoV) coverage. Recognizing the need for more data to propel research in the direction of general-purpose and sensor-agnostic LiDAR data processing algorithms, we have taken the initiative to bridge this gap. Thus, we introduce our novel multi-LiDAR dataset, comprising a spinning LiDAR, two solid-state LiDARs featuring different FoV and scan patterns, and an RGB-D camera combination. This dataset aims to facilitate advancements in UAV tracking and foster the development of robust algorithms capable of handling a diverse range of LiDAR data.

The main contributions of this work are the following:

\begin{enumerate}[label = (\roman*)]
    \item A dataset with data from 3 different LiDAR sensors and an RGB-D camera in both indoor and outdoor environments. This is, to our knowledge, the first diverse dataset in terms of LiDAR sensors for UAV tracking. The dataset includes a spinning LiDARs with 64 (Ouster OS1-64) channels, two different solid-state LiDARs (Livox Mid-360 and Livox Avia) with different scanning patterns and FoVs, and an RGB-D camera (RealSense D435). Given the short range of the point cloud generated by the camera compared to the LiDARs, we only extracted RGB images from it. Low-resolution images with depth, near-infrared, and laser reflectivity data from the Ouster sensor complete the dataset. These are illustrated in Fig.~\ref{fig:hardware}.
    \item The dataset includes sequences with motion capture-based (MOCAP) ground truth in both indoor and outdoor environments. The indoor trajectories exhibit more intricate patterns than the outdoors, while the outdoor sequences were deliberately selected to simulate potential docking and infrastructure inspection scenarios~\cite{seo2018drone} by emphasizing their proximity to a building.
    \item Based on the presented dataset, we provide a baseline comparison with recent LiDAR-based UAV tracking algorithms, benchmarking the performance with different sensors, UAVs of different sizes (from micro-aerial-vehicles to more standard commercial platforms), and algorithms.
\end{enumerate}

Considering the distinctive characteristics of the dataset we have presented, we believe it offers a timely and valuable contribution, complementing the existing datasets that primarily focus on mobile robots indoors or autonomous cars outdoors. The diversity of sensors within our suite paves the way for exploring new challenges within the research community. This dataset serves as a valuable resource to facilitate the development of new algorithms in UAV tracking that effectively address the challenges posed by limited sensor FoV and various scanning modalities. By leveraging this dataset, researchers can make significant strides in enhancing adaptability and robustness in their algorithms, thus advancing the field of LiDAR-based UAV tracking.

In the following sections, we will first provide a brief review of related work on UAV tracking and LiDAR-based sensing in Section II. Then, we will provide an overview of the configuration of the proposed sensor system in detail. Section IV explains the evaluation procedure of available methods on the proposed data set. Finally, Section V concludes the study and introduces suggestions for further work.


\section{Related Work} \label{sec:related_work}

Tracking UAVs using LiDAR technology poses significant challenges due to UAVs' small size, diverse shapes, fast movements, and unpredictability. In the absence of existing datasets dedicated to UAV tracking, we outline various research efforts that have explored innovative methods to enhance the detection and tracking of UAVs utilizing LiDAR technology.

Researchers have investigated different approaches to address the limitations of 3D LiDAR technology and improve UAV detection and tracking. One method involves probabilistic analysis of detections using a rotating turret-mounted LiDAR, enabling a wider FoV coverage with fewer LiDAR beams while continuously tracking only a select number of hits~\cite{dogru2022drone}.

Combining segmentation techniques and a simple object model with temporal information has shown promising results in reducing parametrization efforts and generalizing well in diverse settings~\cite{razlaw2019detection}. Other studies incorporate Euclidean distance clustering and particle filter algorithms to perform UAV detection and tracking~\cite{wang2021study}.

For UAV deployment from ground robots, relative localization between devices is crucial. In previous works, we extensively evaluated the effectiveness of multi-scan integration. Li et al. proposed a multi-modal approach that combines three tracking modalities and integrates multiple scans to adjust point cloud density and size for processing~\cite{qingqing2021adaptive}. Catalano et al. introduced a method that dynamically adjusts the LiDAR frame integration time based on the distance to the
UAV and its speed, fusing two simultaneous scan frequencies using a Kalman filter and Inverse Covariance Intersection for robust and accurate tracking~\cite{catalano2023evaluating, catalano2023uav}.

Cooperative navigation frameworks have been introduced to guide a secondary UAV with unreliable self-localization using a primary UAV equipped with a LiDAR sensor. These frameworks utilize occupancy voxel maps and Kalman Filter-based multi-target tracking techniques\cite{pritzl2022cooperative}.

An innovative ``LiDAR-as-a-camera'' concept fuses images and point cloud data generated by a single LiDAR sensor to track UAVs without prior knowledge. Custom YOLOv5 models trained on panoramic images bring computer vision capabilities to LiDAR technology~\cite{sier2023uav}.

In the context of autonomous UAV landing on a moving ground vehicle, Kim et al. employed a clustering algorithm to identify the UAV within the point cloud data, enabling precise position estimation and the elimination of outliers~\cite{kim2017lidar}.

Departing from traditional track-after-detect approaches, some studies leverage motion information by analyzing 3D details in 360\textdegree\ LiDAR scans and trajectory patterns to classify UAVs and non-UAV objects~\cite{hammer2018potential, hammer2018lidar}.

These research efforts demonstrate the versatility of LiDAR technology for enhancing UAV detection and tracking. However, the absence of a dedicated dataset specific to UAV tracking presents an opportunity for our novel dataset to address this gap and provide valuable resources for advancing UAV tracking algorithms in diverse real-world scenarios.


\section{System Overview}

The data collection's sensor configurations are illustrated in Fig.~\ref{fig:hardware}, while detailed information about each sensor can be found in Table~\ref{table:sensors_details}. As the primary objective of this work is on UAV tracking, the system was mounted on a stationary platform.

\begin{table*}[t]  
 \centering
    \caption{Sensors specification for the presented dataset. Angular resolution is configurable in the OS1-64 (varying the vertical FoV). Livox lidars have a non-repetitive scan pattern that delivers higher angular resolution with longer integration times. Range is based on manufacturer information, with values corresponding to 80\% Lambertian reflectivity and 100 klx sunlight.} 
    \resizebox{\textwidth}{!}{  
    \begin{tabular}{@{}lcccccccc@{}}  
    \toprule
        & IMU & Type & Channels & FoV & Angular Resolution & Range & Freq. & Points   \\
    \midrule   

        \textbf{Ouster OS1-64}      &  ICM-20948 & spinning & 64 & 360\degree×45\degree & V:0.7\degree, H:0.18\degree    & 120\,m       & 20\,Hz  & 1,310,720 pts/s \\ [0.5ex] 

       \textbf{Livox Mid-360}   &  ICM-40609 & solid-state & N/A & 360\degree×59\degree          & N/A                 & 70\,m      & 100\,Hz  & 200,000 pts/s \\ [0.5ex]
        
        \textbf{Livox Avia}      &  BOSCH BMI088 & solid-state & N/A & 70.4\degree×77.2\degree         & N/A                 & 450\,m      & 100\,Hz  & 240,000 pts/s\\ [0.5ex] 
 
        \textbf{RealSense D435}  &  N/A & RGB-D camera & N/A &  69\degree × 42\degree        & V: 1080, H: 1920 at 30 fps       & 10\,m        & 30\,Hz  & -  \\
     \bottomrule
    \end{tabular}
    }
     \label{table:sensors_details}
\end{table*} 

\subsection{Hardware}

The main goal of our sensor system is to gather data from a diverse range of LiDAR sensors, each offering distinct characteristics. These sensors include two innovative low-cost solid-state LiDARs, as well as a 3D spinning LiDAR. Additionally, an RGB-D camera is integrated into the setup. 

Specifically, our data collecting platform consists of a 64-channel Ouster spinning LiDAR (OS1), two Livox solid state LiDAR sensors: Mid-360, featuring a nearly 360\textdegree\ FoV, and Avia, with an almost-circular FoV. The setup is completed with an Intel RealSense D435 RGB-D camera.

\begin{figure}      
     \centering
     \includegraphics[width=0.48\textwidth]{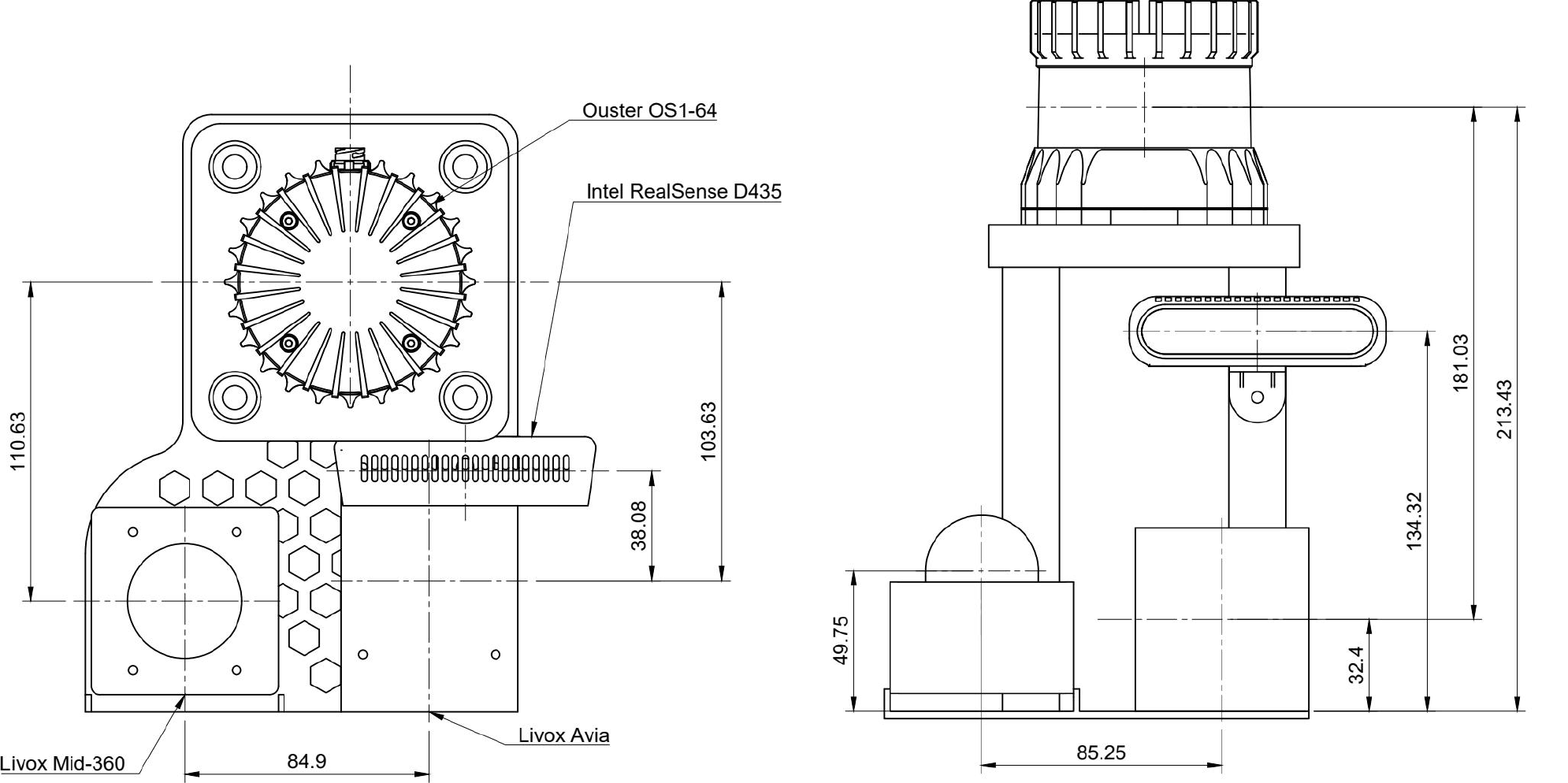}     
    \caption{Data collecting platform, top view (left) and front view (right)} 
    \label{fig:setup_measurements}
\end{figure}

The top and front views depicted in Fig.~\ref{fig:setup_measurements} allow for a comprehensive understanding of the distances, positions, and orientations. 

The LiDARs are linked to a Gigabit Ethernet router, as well as to an onboard computer on the platform. This computer is equipped with an Intel i7-10750h processor, 16\,GB of DDR4 RAM memory, and 1\,TB SSD storage. To maintain a distinct connection from the LiDARs, the Optitrack system is physically attached to the same computer via a separate Ethernet interface. Furthermore, the RealSense D435 camera is connected to a USB 3.2 port for seamless integration and operation.
    
 \begin{figure}[h]
     \centering   
     \includegraphics[width=0.48\textwidth]{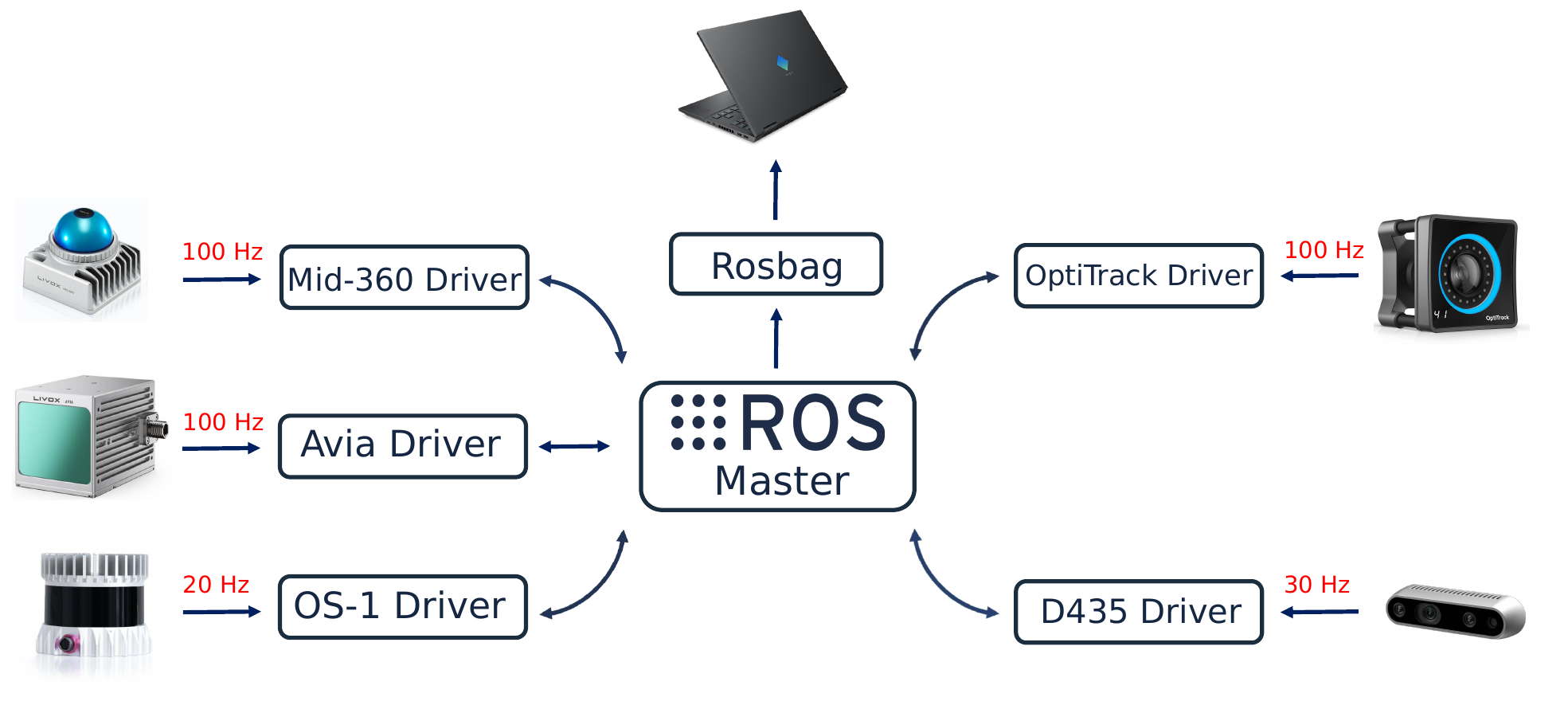}  
     \caption{ROS drivers and data gathering frequency for the different LiDAR sensors used in our platform.}
      \label{fig:software_sensors_config} 
\end{figure}

\subsection{Software}

Our software system is exclusively built upon ROS Noetic, operating on Ubuntu 20.04. The ROS drivers and the publishing frequency of the various sensors are visually represented in Fig.~\ref{fig:software_sensors_config}. To address the absence of hardware signals for sensor data synchronization, as observed in other datasets in the literature~\cite{yan2020eu}, we adopt an approach aimed at minimizing the data synchronization challenge. This involves running all the sensor drivers and data recording programs locally on a high-performance computer. By doing so, in conjunction with the networking equipment, we effectively reduce data transmission latency at both the hardware and software levels. Timestamps are applied at the ROS drivers to ensure synchronization. Additionally, we maintain timestamp consistency across all sensors by leveraging the Precision Time Protocol (PTP), which is specially designed for high-precision time synchronization within local networks.

\subsection{Sensor Calibration}

The extrinsic parameters for the LiDARs were determined using optimization methods similar to those presented in~~\cite{jeong2018complex}. The calibration process was conducted in an indoor office environment with the sensor platform stationary. During calibration, we treated the coordinate system of the Ouster LiDAR sensor as the reference frame. To enhance the level of detail in the environment, ten consecutive frames of point cloud data were integrated from the solid-state LiDARs.

To align the point cloud data from each LiDAR to the reference frame, we employed manual measurements of a set of features in the environment. Subsequently, we utilized the Generalized Iterative Closest Point (GICP) method to iteratively optimize the relative transformation between the reference frame and LiDARs~\cite{segal2009gicp}. This iterative optimization process ensures accurate and precise calibration, enhancing the overall performance of the LiDAR-based system.

Similarly, we determined the extrinsic parameters between the Ouster sensor and the Intel RealSense D435 camera using the depth cloud produced by the latter. We also provide additional stationary data for extrinsic calibration, if a custom calibration is preferred.

The intrinsic parameters of the sensors are given based on factory settings and manufacturer information.

\subsection{Ground Truth}

Accurately generating ground truth data in complex environments is a challenging task, as evidenced by various existing datasets. Many benchmarks rely on GNSS/INS fusion methods to produce ground truth pose data. Nevertheless, in indoor environments, GNSS signals are unavailable, making this approach infeasible. In such indoor settings, MOCAP systems have become widely embraced due to their ability to provide positioning data with millimeter-level accuracy. However, their practicality is limited by the range of the cameras, typically falling within the 10 to 20\,m range. Additionally, the relatively intricate setup required by MOCAP systems has hindered their adoption for outdoor environments.

To address the need for reliable ground-truth data in diverse environments, our present dataset includes MOCAP-based ground-truth data for both indoor and outdoor scenarios. This ensures comprehensive coverage and facilitates robust evaluation in various real-world conditions.

\subsection{Dataset Sequences}

Our dataset is organized into three distinct categories based on the environment and trajectory structure: structured indoor, unstructured indoor, and unstructured outdoor. Each category captures specific movement patterns and characteristics, as follows:

\begin{enumerate}[label = (\roman*)]
    \item Structured Indoor: This subset (HolybroStdn) comprises simple trajectories represented by predefined, systematic patterns, including a circle, a cube, a spiral, and an up and down movement. These structured trajectories are intentionally included to provide standardized, reproducible, and easily interpretable movement patterns. By employing these basic trajectories, ablation studies can be performed on isolated and specific aspects of different methods. This naturally allows for evaluating scenarios where different elements are \textit{decoupled}. The structured indoor trajectories act as a reference point for understanding how well a method performs under well-defined and controlled conditions.
    \item Unstructured: In this subset, trajectories exhibit a more irregular nature, simulating movements that occur both indoors and outdoors without strict adherence to predefined patterns. These trajectories aim to capture the spontaneous and less structured nature of real-world flight scenarios, where a UAV's movements can vary significantly based on the environment and other influencing factors.
\end{enumerate}

A comprehensive list of sequences in each category is shown in  Table~\ref{table:dataset}. 

\begin{table}  
    \centering
    \caption{List of data sequences in our dataset recorded indoor and outdoor} 
    \resizebox{0.90\linewidth}{!}{  
    \begin{tabular}{@{}ccccc@{}}  
    \toprule 
        Sequence       & Description                  & Ground Truth       & Difficulty             \\[0.5ex]
    \midrule
        HolybroStnd01         & Structured(Up/Down)       & MOCAP          & Easy      \\ [0.5ex]
        HolybroStnd02         & Structured(Square)       & MOCAP          & Easy      \\ [0.5ex]
        HolybroStnd03         & Structured(Circle)       & MOCAP          & Easy      \\ [0.5ex]
        HolybroStnd04         & Structured(Spiral)       & MOCAP          & Easy        \\ [0.5ex]
    \midrule   
        Holybro01         & Unstructured, Indoor       & MOCAP          & Easy         \\ [0.5ex]
        Holybro02         & Unstructured, Indoor       & MOCAP          & Easy         \\ [0.5ex]
        Holybro03         & Unstructured, Indoor       & MOCAP          & Easy         \\ [0.5ex]
        Holybro04         & Unstructured, Indoor       & MOCAP          & Medium         \\ [0.5ex]
        Holybro05         & Unstructured, Indoor       & MOCAP          & Medium         \\ [0.5ex]
        HolybroOut01         & Unstructured, Outdoor       & MOCAP          & Medium         \\ [0.5ex]
        HolybroOut02         & Unstructured, Outdoor       & MOCAP          & Medium         \\ [0.5ex]
    \midrule
        Autel01         & Unstructured, Indoor       & MOCAP          & Easy         \\ [0.5ex]
        Autel02         & Unstructured, Indoor       & MOCAP          & Easy         \\ [0.5ex]
        Autel03         & Unstructured, Indoor       & MOCAP          & Easy         \\ [0.5ex]
        Autel04         & Unstructured, Indoor       & MOCAP          & Medium         \\ [0.5ex]
        Autel05         & Unstructured, Indoor       & MOCAP          & Hard         \\ [0.5ex]
        AutelOut01         & Unstructured, Outdoor       & MOCAP          & Hard         \\ [0.5ex]
        AutelOut02         & Unstructured, Outdoor       & MOCAP          & Hard         \\ [0.5ex]
    \midrule
        Tello01         & Unstructured, Indoor       & MOCAP          & Medium      \\ [0.5ex]
        Tello02         & Unstructured, Indoor       & MOCAP          & Medium      \\ [0.5ex]
        Tello03         & Unstructured, Indoor       & MOCAP          & Hard      \\ [0.5ex]
        Tello04         & Unstructured, Indoor       & MOCAP          & Hard        \\ [0.5ex]
        Tello05         & Unstructured, Indoor       & MOCAP          & Hard        \\ [0.5ex]
        TelloOut01         & Unstructured, Outdoor       & MOCAP          & Hard      \\ [0.5ex]
        TelloOut02         & Unstructured, Outdoor       & MOCAP          & Hard      \\ [0.5ex]
    \bottomrule
    \end{tabular}
    }
    \label{table:dataset}
\end{table} 

\begin{figure*}
    \centering   
    \includegraphics[width=\textwidth]{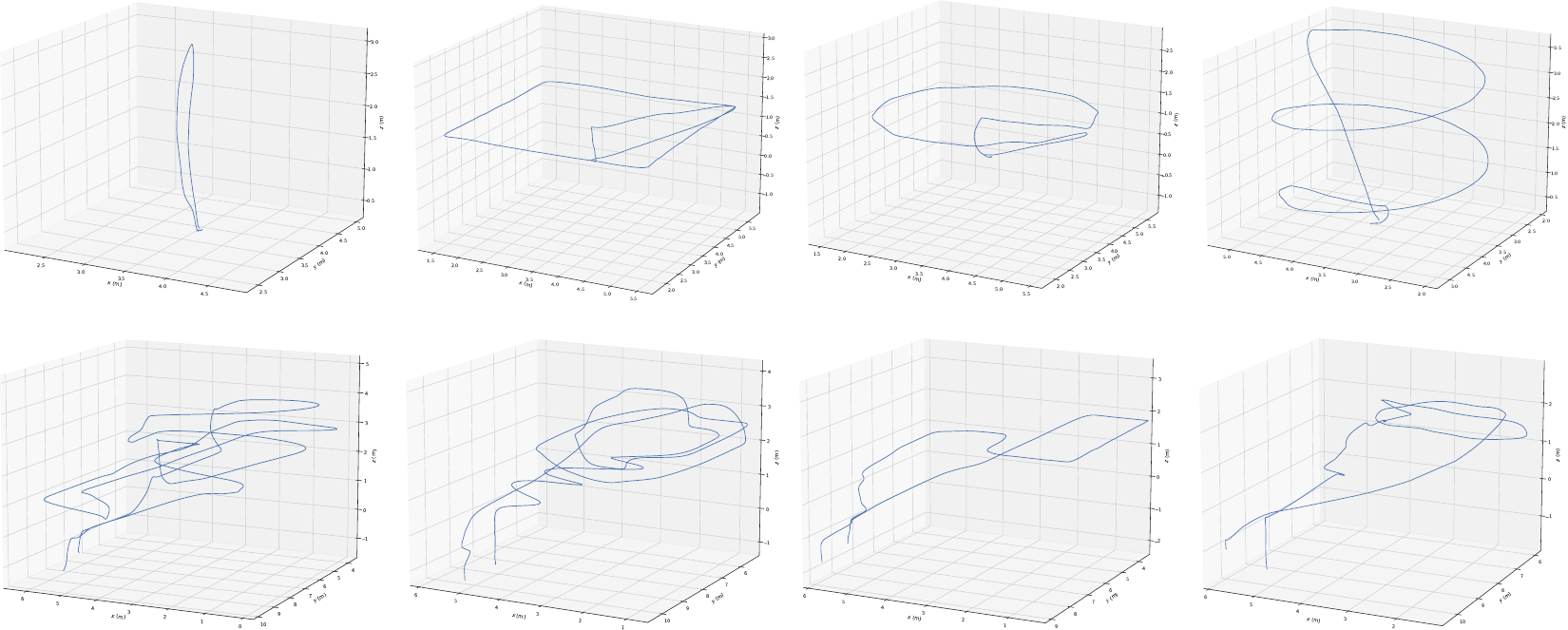}
    \caption{Sample of recorded trajectories. Top row: \textit{HolybroStnd01}, \textit{HolybroStnd02}, \textit{HolybroStnd03}, \textit{HolybroStnd04}. Bottom row: \textit{Holybro03}, \textit{Autel02}, \textit{Autel05}, \textit{Tello04}.}
    \label{fig:traj_samples} 
\end{figure*}

Fig.~\ref{fig:traj_samples} displays a subset of the dataset, represented in three dimensions to enhance the understanding of spatial distances in each direction. Complete visualizations of all recorded trajectories are available on the project's GitHub page.

The data collection for the indoor and outdoor trajectories was conducted in distinct locations to capture a diverse range of environments. The indoor data was gathered in the open area of our lab, providing a controlled and confined setting for UAV movements. On the other hand, the outdoor data was acquired in an open area adjacent to the building, offering a more estensive space for capturing trajectories with greater distances and varied environmental conditions. While the indoor trajectories exhibit both greater complexity and, in some, regularity, the outdoor trajectories emphasize their proximity to the building, simulating potential docking scenarios. 

We qualitatively assessed the complexity of each trajectory accommodating for different UAV capabilities, classifying the trajectories into three levels: easy, medium, and difficult. This categorization is based on the size and speed of the UAV used during data collection. Specifically, we utilized three different UAVs - the Holybro, Autel II, and Tello. As expected, larger UAVs tended to result in easier trajectories, while faster speeds led to higher levels of difficulty. The difficulty levels are chosen with the aim of enabling researchers to evaluate their methods across a spectrum of challenges, ranging from simple to more intricate flight paths, following similar standards in previous datasets~\cite{burri2016euroc}.

To introduce additional challenges, we designed the fourth indoor track for each UAV as a circular path around an obstacle and the fifth indoor track as a loop between two obstacles. These obstacle tracks were specifically included to test the UAVs' maneuvering abilities in constrained spaces, further diversifying the difficulty levels and encouraging the evaluation of methods under more intricate flight conditions.

In our dataset, the indoor trajectories typically span approximately 16\,m, whereas the outdoor trajectories cover larger distances, extending up to 30\,m. This contrast in distances allows researchers to analyze trajectory characteristics in diverse spatial contexts, emphasizing the importance of robust analysis techniques that can adapt to varying scales and complexities.

\subsection{Data Format}

\begin{figure}[]
     \centering
     \includegraphics[width=0.48\textwidth]{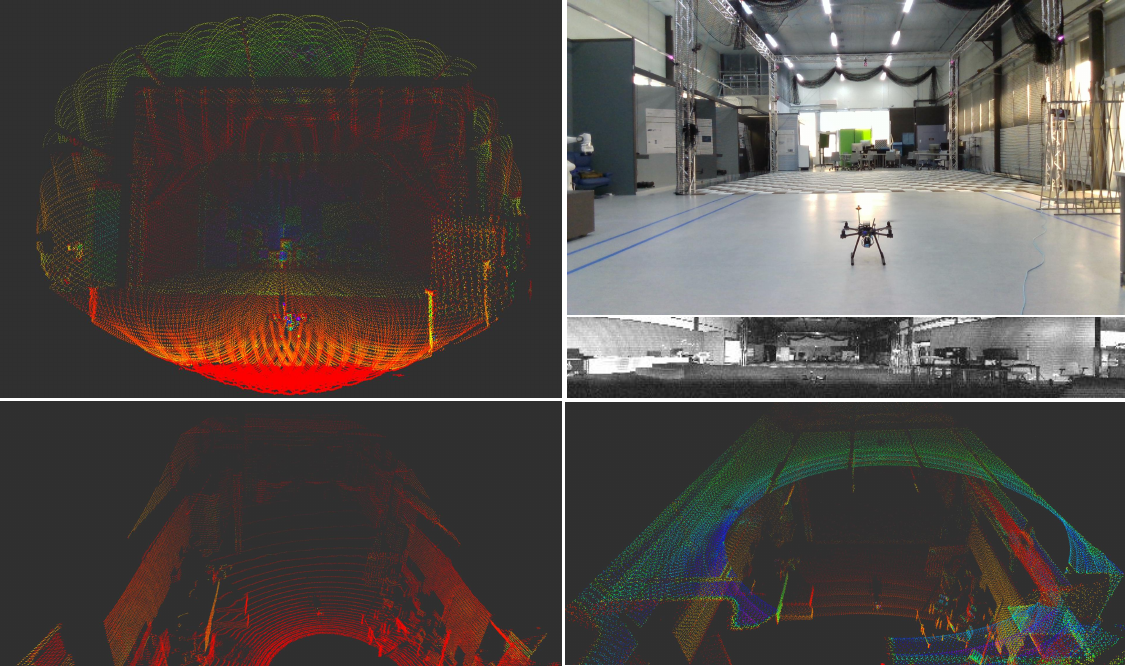}    
     \caption{Subsets of data from the \textit{Holybro01} sequence. The left column shows the LiDAR point cloud from the Avia and the OS-1 as the right column displays the point cloud from the Mid-360 as well as the RGB image from the D435 and the signal image from the OS-1.} 
    \label{fig:data_sample}
\end{figure}

Data collection within the ROS environment makes use of the rosbag format, which has emerged as a standard within the robotics research community. Fig~\ref{fig:data_sample} showcases sampled data frames from a subset of the sensors. Detailed data formats for each type of data included in the dataset are as follows:

\begingroup
\renewcommand{\arraystretch}{1.5}
\begin{table*}[t]
    \centering
    \caption{Position error (RMSE) for the dynamic scan tracking $I_{}^{\text{EKF}}$ and $I_{}^{\text{KF}}$ (N/A when the error diverges because the estimated trajectory is incomplete. Unit: meter)}
    \resizebox{\textwidth}{!}{ 
        \begin{tabular}{@{}lcccccccccc@{}}
            \toprule
            Method & HolybroStnd01 & HolybroStnd02 & HolybroStnd03 & HolybroStnd04 & Holybro01 & Holybro02 & Autel02 & Autel03 & Tello01 & Tello02\\
            \hline
            $I_{\text{Avia}}^{\text{EKF}}$ & 0.0431 & 0.037 & 0.0425 & 0.0526 & \textbf{0.0592} & \textbf{0.0649} & \textbf{0.0984} & 0.183 & \textbf{0.125} & \textbf{0.1182} \\
            
            $I_{\text{Avia}}^{\text{KF}}$ &  \textbf{0.0415} & \textbf{0.0272} & \textbf{0.0389} & \textbf{0.042}  & 0.1155 & N/A & 0.112 & \textbf{0.0395} & N/A & 0.1321 \\

            $I_{\text{Mid-360}}^{\text{EKF}}$ & N/A & N/A & N/A & N/A & N/A & N/A & N/A & N/A & N/A & N/A \\
            
            $I_{\text{Mid-360}}^{\text{KF}}$ &  0.1042 & N/A & N/A & N/A & 0.0673 & N/A & N/A & N/A & N/A & N/A \\

            \bottomrule
        \end{tabular}
    }
    \label{table:evaluation}
\end{table*}

\endgroup
\begin{table}  
    \centering
    \caption{Percentage of successfully estimated trajectories for he dynamic scan tracking $I_{}^{\text{EKF}}$ and $I_{}^{\text{KF}}$ on the selected data subset (N/A when the method was not designed for that type of LiDAR)} 
    \begin{tabular}{@{}ccccc@{}}  
    \toprule 
        LiDAR & $I_{}^{\text{EKF}}$ & $I_{}^{\text{KF}}$ \\[0.5ex]
    \midrule
        Livox Avia & 100\% & 80\% \\ [0.5ex]
        Livox Mid-360 & 0\% &  20\% \\ [0.5ex]
        Ouster & N/A & N/A \\ [0.5ex]
    \bottomrule
    \end{tabular}
    \label{table:evaluation_percentage}
\end{table} 

\begin{enumerate}[label = (\roman*)]
    \item \textbf{Point cloud data from spinning LiDAR (OS1-64)} is recorded as \textit{sensor\_msgs::PointCloud}. Each point in the point cloud contains four values $(x, y , z, I)$, representing local Cartesian coordinates $(x,y,z)$, and the measured laser reflectance $(I)$.
    \item \textbf{Point cloud data from solid-state LiDARs, Avia, and Mid-360}, employs Livox's custom data format named \textit{livox\_ros\_driver/CustomMsg} in the rosbags. This custom message includes a base time and an offset time relative to the base time for each point. This approach compensates for the non-repetitive pattern inherent to solid-state LiDARs and allows for de-skewing of the point cloud data, addressing distortions caused by the sensor's egomotion. We have retained this message type for algorithms involving point cloud deskewing and related research. However, to facilitate visualization in tools like Rviz and compatibility with standard LiDAR processing algorithms relying on ROS messages, we provide format conversion tools to transform the Livox custom message data to the standard ROS message type, \textit{sensor\_msgs::PointCloud}. The converted points now hold five values $(x, y, z, I, C)$, where $x, y, z$ represent local Cartesian coordinates, $I$ is the intensity of the point, and $C$ incorporates the line (integer part) and point timestamp (decimal part).
    \item \textbf{Images from RGB camera}. The RealSense D435 camera publishes RGB images at 1920×1080 resolution and a 30\,Hz frequency. The message type is $sensor\_msgs::Image$.
    \item \textbf{Images from the high-resolution spinning LiDAR (OS1-64)}  consist of fixed-resolution range images, near-infrared images captured by the laser sensor, and signal images. Each pixel in these images represents the distance from the sensor origin to the point, the captured light's strength, and the object's reflectivity, respectively. These images are published at a frequency of 10\,Hz and have 16 bits per pixel with a linear photo response. The standard ROS message type, \textit{sensor\_msgs/Image}, is used for these image data.
    \item \textbf{Inertial data} is available from both spinning and solid-state LiDARs, featuring three built-in 6-axis IMU sensors with a 3-axis gyroscope and a 3-axis accelerometer. The IMU data is published at a frequency of 100\,Hz for the spinning LiDAR and 200\,Hz for the solid-state LiDAR. The standard ROS message type, \textit{sensor\_msgs::Imu}, is employed for IMU data in the rosbags.    
    \item \textbf{Ground truth data} is derived from the MOCAP system and is included in the rosbags as \textit{geometry\_msgs::PoseStamped} messages. These data are obtained from the computer connected to the OptiTrack cameras via a VRPN connection, providing precise ground truth information.    
\end{enumerate}


\section{Experimental Evaluation}

This section covers the characterization of the different tracking approaches for the dataset sequences.

\subsection{Metrics}

First, to quantify the disparity between the different LiDAR sensors and the external position system estimates, we computed the error by taking the difference between the position estimates obtained from both systems for two distinct positions and orientations of the target. This analysis revealed a Root Mean Squared Error (RMSE) of 0.0143\,m.

To quantitatively evaluate the tracking performance, we employed the Root Mean Squared Error (RMSE) metric, and the summarized outcomes are presented in Table~\ref{table:evaluation}. Additionally, we provide an overall assessment of each method's performance based on the percentage of successfully estimated trajectories, as shown in Table~\ref{table:evaluation_percentage}.

\subsection{Experiments}

As part of our dataset, we provide an evaluation of current UAV tracking methods on several sequences. The objective is to compare the performance of different methods to provide a baseline for future research. Throughout this section, we discuss the best methods for different types of LiDAR sensors and environments.

From our analysis, one of the key findings is the pressing need for methods that can enhance UAV tracking with sparse LiDAR data, regardless of the scanning pattern.

Regarding the dynamic tracking method~\cite{catalano2023uav} ($I_{}^{\text{EKF}}$ and $I_{}^{\text{KF}}$), initially designed for the Livox Horizon LiDAR, we observed that it exhibits strong generalization capabilities and performs effectively on the Livox Avia.

Among the tested methods, we observed that tracking using the dense solid-state Livox Avia LiDAR yields superior results compared to the sparser Mid-360. Specifically, the $I_{}^{\text{EKF}}$ method successfully estimated 100\% of the selected trajectories, while the $I_{}^{\text{KF}}$ method achieved better results on more standard patterns, such as \textit{HolybroStdn}. However, it's worth noting that the dynamic tracking method, being designed for dense solid-state LiDARs, exhibited poor performance on the Mid-360, regardless of the trajectory type and UAV used.


In summary, our evaluation highlights the need for improved UAV tracking methods, especially for sparse LiDAR data, and demonstrates the varying performance of different methods based on the type of LiDAR sensor and scanning environment. These findings lay the groundwork for future research and development in UAV tracking techniques.

\section{Conclusion}\label{sec:conclusion}

We introduced a novel dataset collected through a multi-LiDAR sensor system, including both indoor and outdoor environments. The dataset comprises diverse LiDAR types with varying resolutions and scan patterns, along with an RGB camera. We also include UAVs with variable sizes common to urban areas. This diverse range of sensors and UAVs provides a unique opportunity for future research on general-purpose algorithms, as our analysis reveals distinct performance variations among different algorithms based on the type of LiDAR used. Consequently, there exists a substantial opportunity to develop more robust LiDAR-based UAV tracking algorithms.

To facilitate algorithm analysis, we have included ground truth data for both indoor and outdoor settings. The dataset's distinctive characteristics, encompassing a wide array of data and environmental conditions, distinguish it from existing literature. This paper thereby aims at establishing a solid foundation for benchmarking and performing quantitative comparisons between existing and forthcoming LiDAR-based UAV tracking algorithms. This dataset promises to advance the field of UAV tracking, opening up new avenues for cutting-edge research and development.




\bibliographystyle{unsrt}
\bibliography{bibliography}

\end{document}